\documentclass{article}

\usepackage[preprint]{neurips_2023}
\usepackage[numbers]{natbib}
\usepackage{graphicx}

\usepackage[utf8]{inputenc} 
\usepackage[T1]{fontenc}    
\usepackage[colorlinks,
    colorlinks=true, 
    linkcolor=blue, 
    filecolor=blue, 
    citecolor=blue,       
    urlcolor=blue]{hyperref}       
\usepackage{url}            
\usepackage{booktabs}       
\usepackage{amsfonts}       
\usepackage{amsmath}
\usepackage{amsthm}
\usepackage{amssymb}
\usepackage{nicefrac}       
\usepackage{microtype}      
\usepackage{xcolor}         
\usepackage[font=small,labelfont=bf]{caption}
\usepackage{enumitem}
\usepackage{wrapfig}
\usepackage{listings}
\usepackage{caption}

\usepackage[normalem]{ulem}
\usepackage{xspace}
\usepackage{float}
\usepackage{tabularx}
\usepackage[normalem]{ulem}
\useunder{\uline}{\ul}{}

\def\llama{Llama\xspace}
\def\mistral{Mistral~7B\xspace}

\def\mistralchat{Mistral~7B~--~Instruct\xspace}

\newcommand{\bt}{\textasciigrave}

\title{\mistral}

\author{%
Albert Q. Jiang, Alexandre Sablayrolles, Arthur Mensch, Chris Bamford, \\
\textbf{Devendra Singh Chaplot, Diego de las Casas, Florian Bressand, Gianna Lengyel,}\\
\textbf{Guillaume Lample, Lucile Saulnier, Lélio Renard Lavaud, Marie-Anne Lachaux,} \\
\textbf{Pierre Stock, Teven Le Scao, Thibaut Lavril, Thomas Wang, Timothée Lacroix,}\\
\textbf{William El Sayed}\\
} 

\begin{document}

\maketitle

\begin{center}
\vspace{-30pt}
\centering
\includegraphics[width=0.8\linewidth,keepaspectratio]{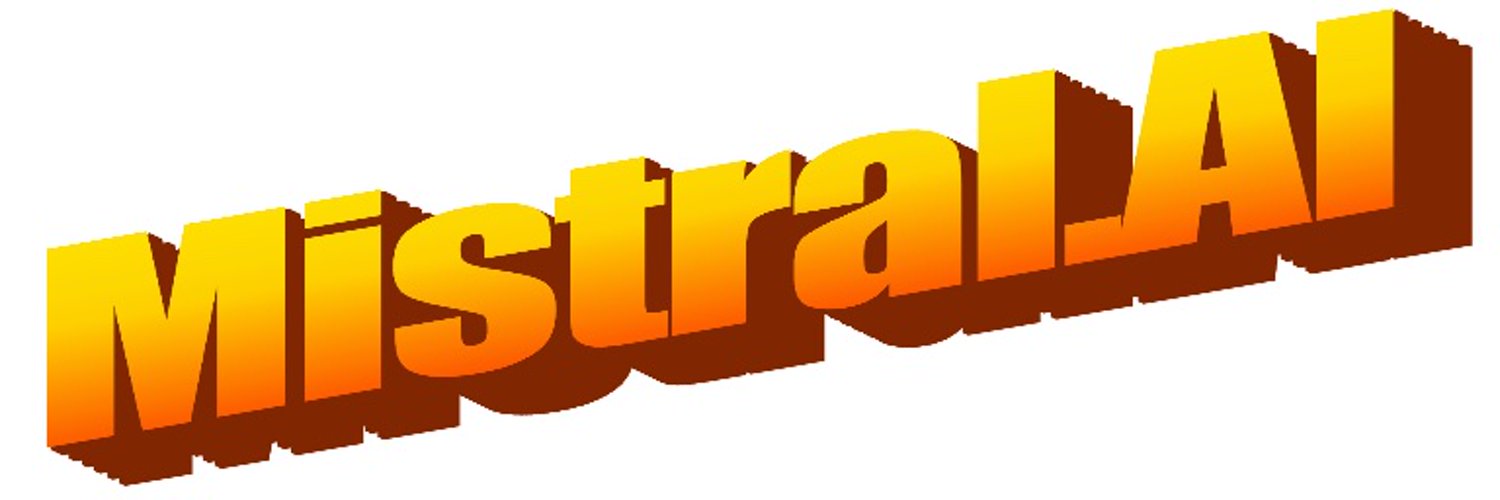}
\end{center}

\begin{abstract}
We introduce \mistral, a 7--billion-parameter language model engineered for superior performance and efficiency.
\mistral outperforms the best open 13B model (Llama 2) across all evaluated benchmarks, and the best released 34B model (Llama 1) in reasoning, mathematics, and code generation.
Our model leverages grouped-query attention (GQA) for faster inference, coupled with sliding window attention (SWA) to effectively handle sequences of arbitrary length with a reduced inference cost.
We also provide a model fine-tuned to follow instructions, \mistralchat, that surpasses \llama~2~13B~--~chat model both on human and automated benchmarks.
Our models are released under the Apache 2.0 license.\\
\textbf{Code:} \url{https://github.com/mistralai/mistral-src} \\
\textbf{Webpage:} \url{https://mistral.ai/news/announcing-mistral-7b/}
\end{abstract}

\section{Introduction}

\looseness=-1 In the rapidly evolving domain of Natural Language Processing (NLP), the race towards higher model performance often necessitates an escalation in model size.
However, this scaling tends to increase computational costs and inference latency, thereby raising barriers to deployment in practical, real-world scenarios.
In this context, the search for balanced models delivering both high-level performance and efficiency becomes critically essential.
Our model, \mistral, demonstrates that a carefully designed language model can deliver high performance while maintaining an efficient inference.
\mistral outperforms the previous best 13B model (Llama 2, \cite{touvron2023llama2}) across all tested benchmarks, and surpasses the best 34B model (LLaMa 34B,~\cite{touvron2023llama}) in mathematics and code generation.
Furthermore, \mistral approaches the coding performance of Code-\llama 7B~\cite{roziere2023code}, without sacrificing performance on non-code related benchmarks.

\mistral leverages grouped-query attention (GQA)~\cite{ainslie2023gqa}, and sliding window attention (SWA)~\cite{child2019generating,beltagy2020longformer}. GQA significantly accelerates the inference speed, and also reduces the memory requirement during decoding, allowing for higher batch sizes hence higher throughput, a crucial factor for real-time applications.
In addition, SWA is designed to handle longer sequences more effectively at a reduced computational cost, thereby alleviating a common limitation in LLMs. These attention mechanisms collectively contribute to the enhanced performance and efficiency of \mistral.

\mistral is released under the Apache 2.0 license.
This release is accompanied by a reference implementation\footnote{\url{https://github.com/mistralai/mistral-src}} facilitating easy deployment either locally or on cloud platforms such as AWS, GCP, or Azure using the vLLM~\cite{kwon2023efficient} inference server and SkyPilot~\footnote{\url{https://github.com/skypilot-org/skypilot}}.
Integration with Hugging Face~\footnote{\url{https://huggingface.co/mistralai}} is also streamlined for easier integration.
Moreover, \mistral is crafted for ease of fine-tuning across a myriad of tasks.
As a demonstration of its adaptability and superior performance, we present a chat model fine-tuned from \mistral that significantly outperforms the \llama 2 13B -- Chat model.

\looseness=-1 \mistral takes a significant step in balancing the goals of getting high performance while keeping large language models efficient.
Through our work, our aim is to help the community create more affordable, efficient, and high-performing language models that can be used in a wide range of real-world applications.

\section{Architectural details}

\begin{figure}[h]
\centering
\includegraphics[width=0.99\linewidth,height=\textheight,keepaspectratio]{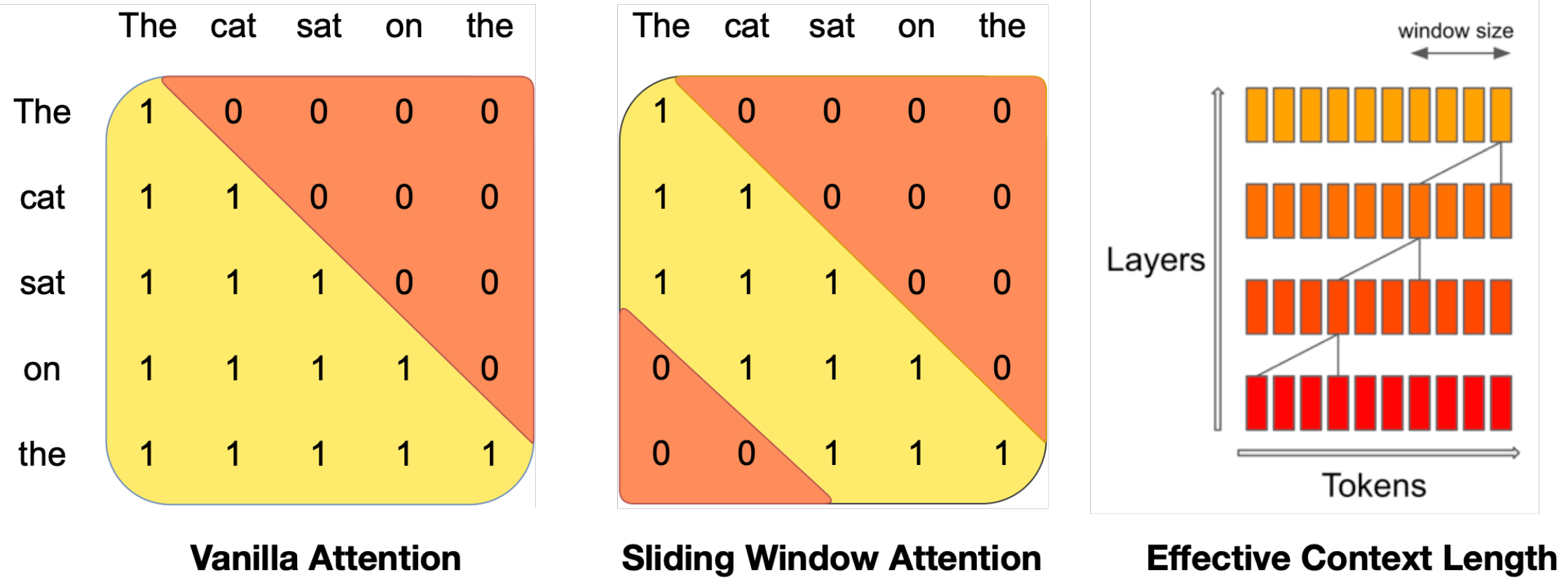}
\caption{\small \textbf{Sliding Window Attention.} The number of operations in vanilla attention is quadratic in the sequence length, and the memory increases linearly with the number of tokens. At inference time, this incurs higher latency and smaller throughput due to reduced cache availability. To alleviate this issue, we use sliding window attention: each token can attend to at most $W$ tokens from the previous layer (here, $W=3$). Note that tokens outside the sliding window still influence next word prediction. At each attention layer, information can move forward by $W$ tokens. Hence, after $k$ attention layers, information can move forward by up to $k \times W$ tokens.}
\label{fig:swa}
\end{figure}

\begin{wrapfigure}{r}{0.275\textwidth}
\center
\small
\vspace{-15pt}
\begin{tabular}{lr}
\toprule
\textbf{Parameter}  & \textbf{Value} \\ \midrule
\texttt{dim}             & $4096$           \\
\texttt{n\_layers}       & $32$             \\
\texttt{head\_dim}       & $128$            \\
\texttt{hidden\_dim}     & $14336$          \\
\texttt{n\_heads}        & $32$             \\
\texttt{n\_kv\_heads}    & $8$              \\
\texttt{window\_size}    & $4096$           \\
\texttt{context\_len}    & $8192$           \\
\texttt{vocab\_size}     & $32000$          \\ \bottomrule
\end{tabular}
\captionof{table}{\small \textbf{Model architecture.}}
\label{tab:param}
\vspace{-8pt}
\end{wrapfigure}

\mistral is based on a transformer architecture~\cite{vaswani2017attention}. The main parameters of the architecture are summarized in Table~\ref{tab:param}. Compared to \llama, it introduces a few changes that we summarize below.

\looseness=-1 \textbf{Sliding Window Attention.} SWA exploits the stacked layers of a transformer to attend information beyond the window size $W$.
The hidden state in position $i$ of the layer $k$, $h_i$, attends to all hidden states from the previous layer with positions between $i-W$ and $i$.
Recursively, $h_i$ can access tokens from the input layer at a distance of up to $W \times k$ tokens, as illustrated in Figure~\ref{fig:swa}.
At the last layer, using a window size of $W=4096$, we have a theoretical attention span of approximately $131K$ tokens.
In practice, for a sequence length of 16K and $W=4096$, changes made to FlashAttention~\cite{dao2022flashattention} and xFormers~\cite{xFormers2022} yield a 2x speed improvement over a vanilla attention baseline.

\begin{figure*}

\makebox[\textwidth][c]{\includegraphics[width=1.0\linewidth,height=\textheight,keepaspectratio]{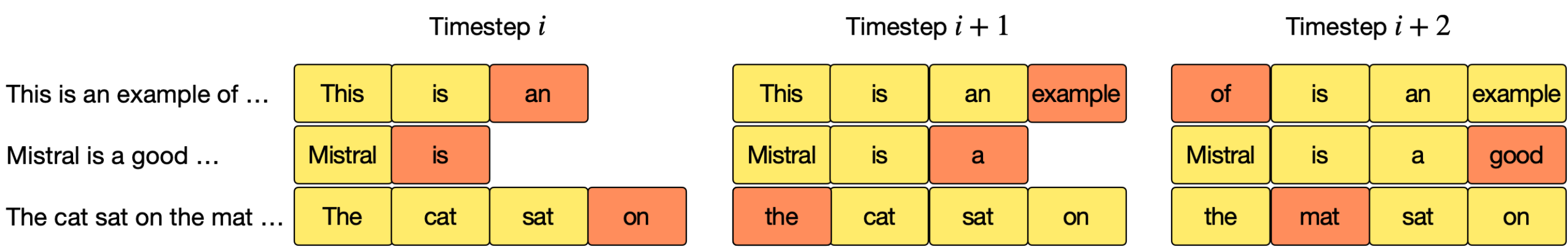}}
\caption{\small \textbf{Rolling buffer cache.} The cache has a fixed size of $W=4$. Keys and values for position $i$ are stored in position $i \bmod W$ of the cache. When the position $i$ is larger than $W$, past values in the cache are overwritten.
The hidden state corresponding to the latest generated tokens are colored in orange.
}
\label{fig:cache}
\end{figure*}

\looseness=-1 \textbf{Rolling Buffer Cache.} A fixed attention span means that we can limit our cache size using a rolling buffer cache.
The cache has a fixed size of $W$, and the keys and values for the timestep $i$ are stored in position $i \bmod W$ of the cache. As a result, when the position $i$ is larger than $W$, past values in the cache are overwritten, and the size of the cache stops increasing. We provide an illustration in Figure~\ref{fig:cache} for $W=3$.
On a sequence length of 32k tokens, this reduces the cache memory usage by 8x, without impacting the model quality.

\begin{figure*}[h]
\centering
\includegraphics[width=0.7\linewidth]{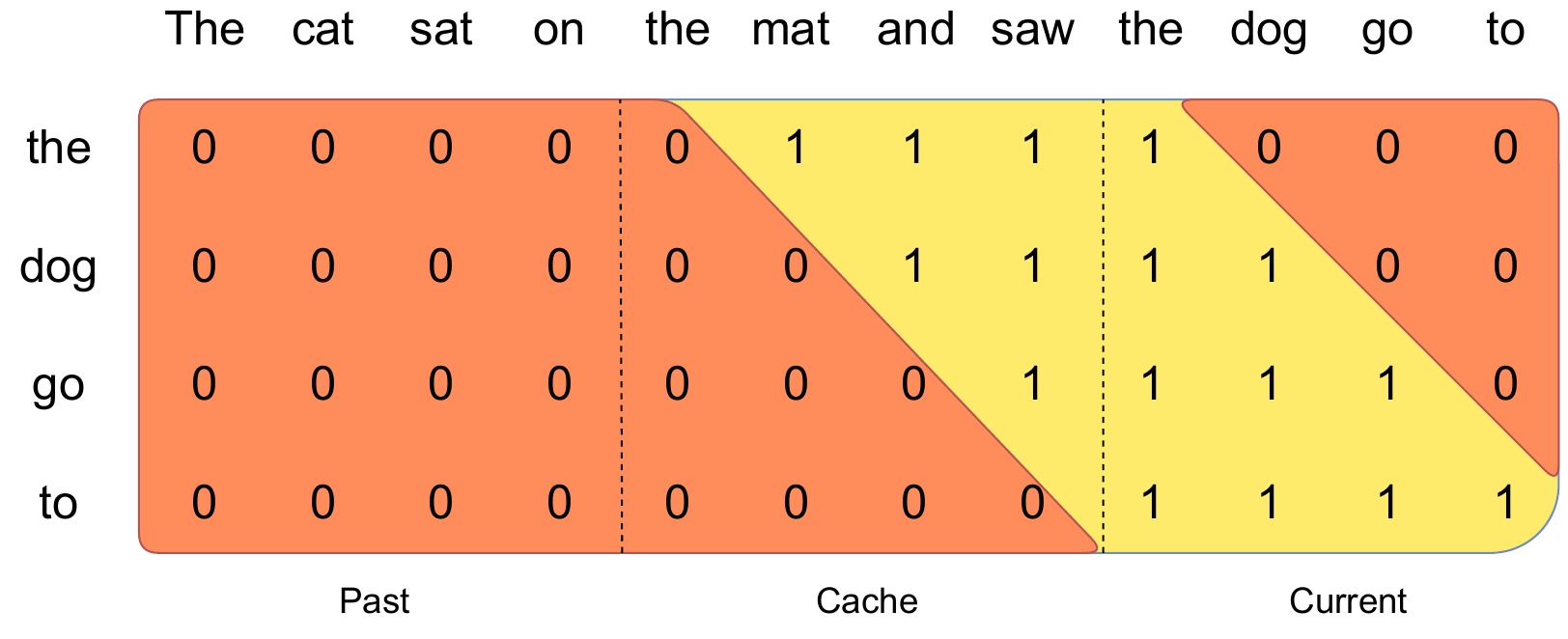} 
\caption{
\small
\textbf{Pre-fill and chunking.}
During pre-fill of the cache, long sequences are chunked to limit memory usage.
We process a sequence in three chunks, ``The cat sat on'', ``the mat and saw'', ``the dog go to''. 
The figure shows what happens for the third chunk (``the dog go to''): it attends itself using a causal mask (rightmost block), attends the cache using a sliding window (center block), and does not attend to past tokens as they are outside of the sliding window (left block).
}
\label{fig:chunking}
\vspace{0.1in}
\end{figure*}

\looseness=-1 \textbf{Pre-fill and Chunking.} When generating a sequence, we need to predict tokens one-by-one, as each token is conditioned on the previous ones. However, the prompt is known in advance, and we can pre-fill the ($k$, $v$) cache with the prompt. If the prompt is very large, we can chunk it into smaller pieces, and pre-fill the cache with each chunk. For this purpose, we can select the window size as our chunk size. 
For each chunk, we thus need to compute the attention over the cache and over the chunk. 
Figure~\ref{fig:chunking} shows how the attention mask works over both the cache and the chunk.

\section{Results}

We compare \mistral to \llama, and re-run all benchmarks with our own evaluation pipeline for fair comparison.
We measure performance on a wide variety of tasks categorized as follow:

\begin{itemize}[leftmargin=10pt]
\item \textbf{Commonsense Reasoning (0-shot):} Hellaswag~\cite{zellers2019hellaswag}, Winogrande~\cite{sakaguchi2021winogrande}, PIQA~\cite{bisk2020piqa}, SIQA~\cite{sap2019socialiqa}, OpenbookQA~\cite{mihaylov2018can}, ARC-Easy, ARC-Challenge~\cite{clark2018think}, CommonsenseQA~\cite{talmor2018commonsenseqa}
\item \textbf{World Knowledge (5-shot):} NaturalQuestions~\cite{kwiatkowski2019natural}, TriviaQA~\cite{joshi2017triviaqa}
\item \textbf{Reading Comprehension (0-shot):} BoolQ~\cite{clark2019boolq}, QuAC~\cite{choi2018quac}
\item \textbf{Math:} GSM8K~\cite{cobbe2021training} (8-shot) with maj@8 and MATH~\cite{hendrycks2021measuring} (4-shot) with maj@4
\item \textbf{Code:} Humaneval~\cite{chen2021evaluating} (0-shot) and MBPP~\cite{austin2021program} (3-shot)
\item \textbf{Popular aggregated results:} MMLU~\cite{hendrycks2020measuring} (5-shot), BBH~\cite{suzgun2022challenging} (3-shot), and AGI Eval~\cite{zhong2023agieval} (3-5-shot, English multiple-choice questions only)
\end{itemize}

\begin{figure*}
\centering
\includegraphics[width=0.99\linewidth,height=\textheight,keepaspectratio]{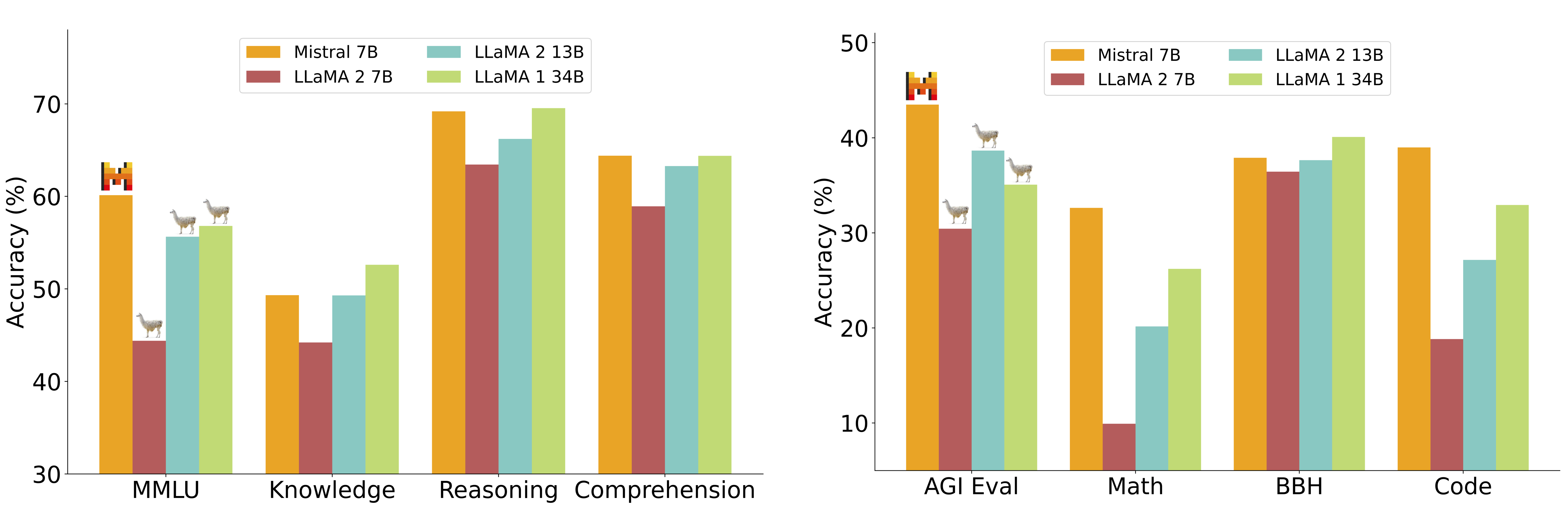}
\caption{\small \textbf{Performance of \mistral and different \llama models on a wide range of benchmarks}. All models were re-evaluated on all metrics with our evaluation pipeline for accurate comparison. \mistral significantly outperforms \llama 2 7B and \llama 2 13B on all benchmarks. It is also vastly superior to \llama 1 34B in mathematics, code generation, and reasoning benchmarks.}
\label{fig:bars}
\end{figure*}

Detailed results for \mistral, \llama 2 7B/13B, and Code-\llama 7B are reported in Table~\ref{tab:results}.
Figure~\ref{fig:bars} compares the performance of \mistral with \llama 2 7B/13B, and \llama 1 34B\footnote{Since \llama 2 34B was not open-sourced, we report results for \llama 1 34B.} in different categories.
\mistral surpasses \llama 2 13B across all metrics, and outperforms \llama~1~34B on most benchmarks.
In particular, \mistral displays a superior performance in code, mathematics, and reasoning benchmarks.

\begin{figure*}
\centering
\includegraphics[width=0.7\linewidth,height=\textheight,keepaspectratio]{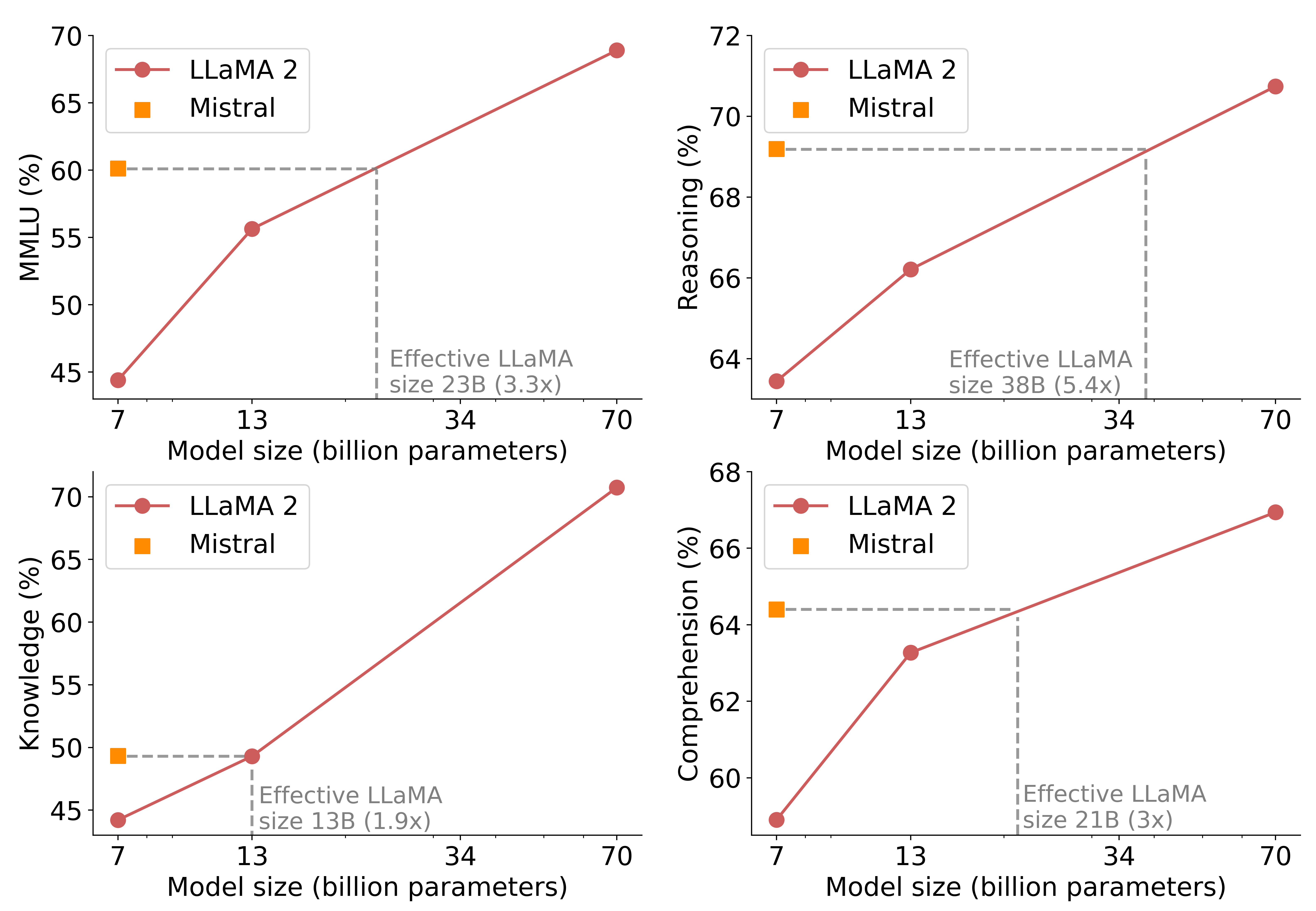}
\caption{\small \textbf{Results on MMLU, commonsense reasoning, world knowledge and reading comprehension for \mistral and \llama 2 (7B/13B/70B)}. \mistral largely outperforms \llama 2 13B on all evaluations, except on knowledge benchmarks, where it is on par (this is likely due to its limited parameter count, which limits the amount of knowledge it can compress).}
\label{fig:size}
\end{figure*}

\textbf{Size and Efficiency.} We computed ``equivalent model sizes'' of the \llama 2 family, aiming to understand \mistral models' efficiency in the cost-performance spectrum (see Figure~\ref{fig:size}). When evaluated on reasoning, comprehension, and STEM reasoning (specifically MMLU), \mistral mirrored performance that one might expect from a \llama 2 model with more than 3x its size. On the Knowledge benchmarks, \mistral's performance achieves a lower compression rate of 1.9x, which is likely due to its limited parameter count that restricts the amount of knowledge it can store.

\textbf{Evaluation Differences.} On some benchmarks, there are some differences between our evaluation protocol and the one reported in the \llama 2 paper: 1) on MBPP, we use the hand-verified subset 2) on TriviaQA, we do not provide Wikipedia contexts.

\setlength{\tabcolsep}{1.8pt}
\begin{table}
{\scriptsize
\centering
\begin{tabular}{@{}lccccccccccccc@{}}
\toprule
Model         & Modality   & MMLU            & HellaSwag       & WinoG      & PIQA            & Arc-e           & Arc-c           & NQ              & TriviaQA        & HumanEval       & MBPP            & MATH            & GSM8K           \\ \midrule
LLaMA 2 7B    & Pretrained & 44.4\%          & 77.1\%          & 69.5\%          & 77.9\%          & 68.7\%          & 43.2\%          & 24.7\%          & 63.8\%          & 11.6\%          & 26.1\%          & 3.9\%           & 16.0\%          \\
LLaMA 2 13B   & Pretrained & 55.6\%          & \textbf{80.7\%} & 72.9\%          & 80.8\%          & 75.2\%          & 48.8\%          & \textbf{29.0\%} & \textbf{69.6\%} & 18.9\%          & 35.4\%          & 6.0\%           & 34.3\%          \\ \midrule
Code-\llama 7B & Finetuned  & 36.9\%          & 62.9\%          & 62.3\%          & 72.8\%          & 59.4\%          & 34.5\%          & 11.0\%          & 34.9\%          & \textbf{31.1\%} & \textbf{52.5\%} & 5.2\%           & 20.8\%          \\ \midrule
\mistral    & Pretrained & \textbf{60.1\%} & \textbf{81.3\%} & \textbf{75.3\%} & \textbf{83.0\%} & \textbf{80.0\%} & \textbf{55.5\%} & \textbf{28.8\%} & \textbf{69.9\%} & \textbf{30.5\%} & 47.5\%          & \textbf{13.1\%} & \textbf{52.2\%} \\ \bottomrule
\end{tabular}
}
\vspace{4pt}
\caption{\small \textbf{Comparison of \mistral with \llama.} \mistral outperforms \llama 2 13B on all metrics, and approaches the code performance of Code-\llama 7B without sacrificing performance on non-code benchmarks.}
\label{tab:results}
\end{table}

\begin{wrapfigure}{r}{0.48\textwidth}
{\centering
\small
\vspace{-13pt}
\setlength{\tabcolsep}{2pt}
\begin{tabular}{@{}lcl@{}}
\toprule
\textbf{Model}      & \textbf{\begin{tabular}[c]{@{}c@{}}Chatbot Arena\\  ELO Rating\end{tabular}} & \textbf{MT Bench} \\ \midrule
WizardLM 13B v1.2   & 1047                                                                         & 7.2               \\
\textbf{Mistral 7B Instruct} & \textbf{1031}                                                       & \textbf{6.84 +/- 0.07}    \\
Llama 2 13B Chat    & 1012                                                                         & 6.65              \\
Vicuna 13B           & 1041                                                                         & 6.57              \\
Llama 2 7B Chat     & 985                                                                          & 6.27              \\
Vicuna 7B           & 997                                                                          & 6.17              \\
Alpaca 13B          & 914                                                                          & 4.53              \\ \bottomrule
\end{tabular}
\vspace{-3pt}
\captionof{table}{\small \textbf{Comparison of Chat models.} \mistralchat outperforms all 7B models on MT-Bench, and is comparable to 13B -- Chat models.}
\label{tab:results_finetuning}
}
\vspace{-10pt}
\end{wrapfigure}

\section{Instruction Finetuning}

\looseness=-1 To evaluate the generalization capabilities of \mistral, we fine-tuned it on instruction datasets publicly available on the Hugging Face repository.
No proprietary data or training tricks were utilized: \mistralchat model is a simple and preliminary demonstration that the base model can easily be fine-tuned to achieve good performance.
In Table~\ref{tab:results_finetuning}, we observe that the resulting model, \mistralchat, exhibits superior performance compared to all 7B models on MT-Bench, and is comparable to 13B -- Chat models.
An independent human evaluation was conducted on \url{https://llmboxing.com/leaderboard}.

\begin{figure*}
\centering
\includegraphics[width=1.0\linewidth,keepaspectratio]{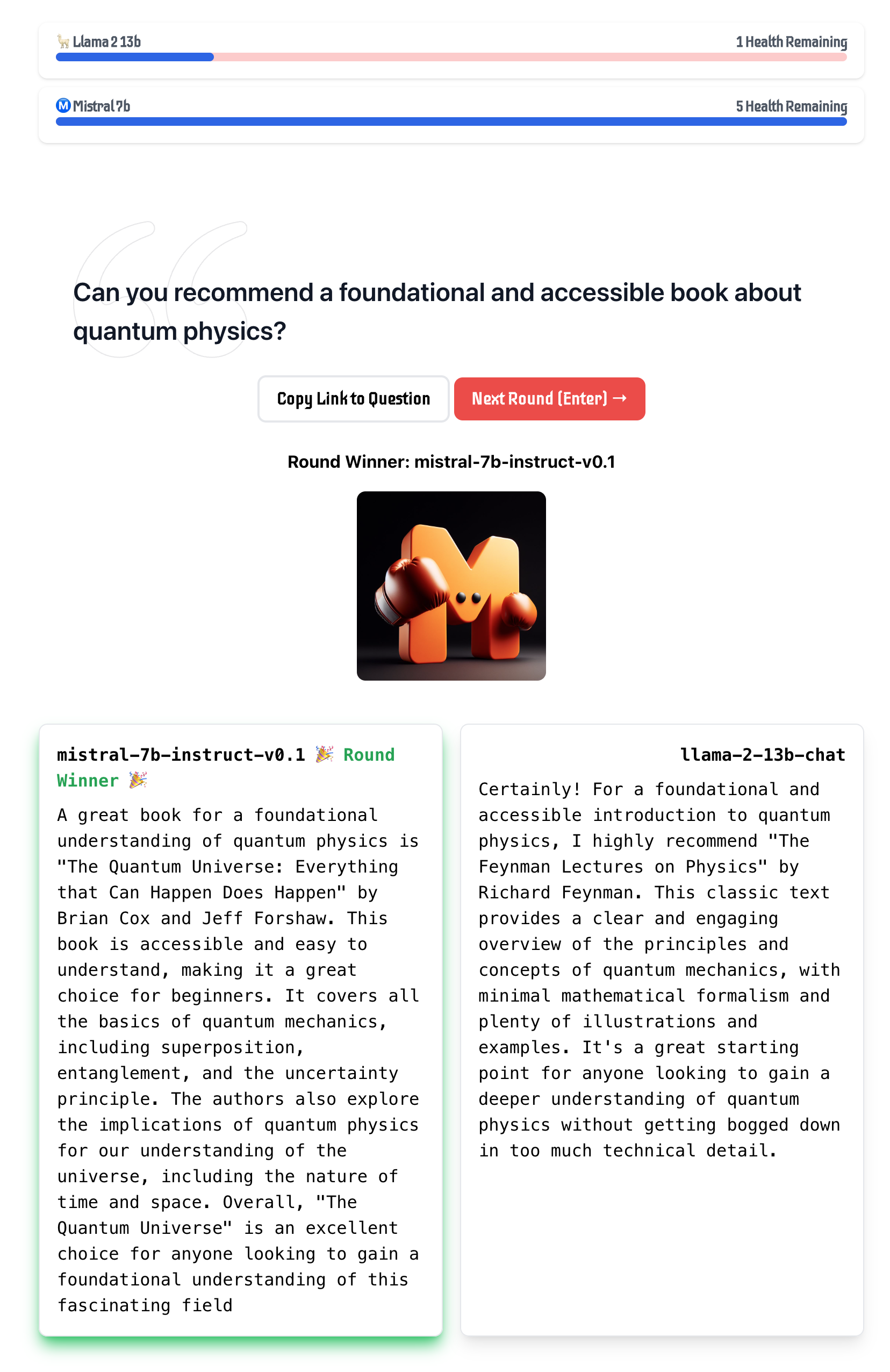}
\vspace{-10pt}
\caption{\small \textbf{Human evaluation of \mistralchat vs \llama 2~13B~--~Chat Example.} An example of human evaluation from \url{llmboxing.com}. The question asks for recommendations of books in quantum physics. \llama 2 13B -- Chat recommends a general physics book, while \mistralchat recommends a more relevant book on quantum physics and describes in the contents in more detail.}
\label{fig:humanevalquestion}
\end{figure*}

In this evaluation, participants were provided with a set of questions along with anonymous responses from two models and were asked to select their preferred response, as illustrated in Figure~\ref{fig:humanevalquestion}.
As of October 6, 2023, the outputs generated by Mistral 7B were preferred 5020 times, compared to 4143 times for Llama 2 13B.

\section{Adding guardrails for front-facing applications}

\looseness=-1 The ability to enforce guardrails when it comes to AI generation is important for front-facing applications.
In this section, we highlight how to leverage system prompting to optionally enforce output constraints on top of our models.
Additionally, we showcase the ability of \mistral to perform fine-grained content moderation, which can be useful to enforce quality content in applications.

\subsection{System prompt to enforce guardrails}

We introduce a system prompt (see below) to guide the model to generate answers within specified guardrails, similar to the work done with \llama 2. Using this prompt allows the user to move on the Pareto front of model utility / guardrails enforcement, as indicated in Table~\ref{tab:guardrails}.

\noindent\fbox{%
    \parbox{0.98\textwidth}{%
        Always assist with care, respect, and truth. Respond with utmost utility yet securely. Avoid harmful, unethical, prejudiced, or negative content. Ensure replies promote fairness and positivity.
    }%
}

\setlength{\tabcolsep}{6pt}
\begin{wrapfigure}{r}{0.38\textwidth}
{\centering
\small
\vspace{-10pt}
\begin{tabular}{@{}cc@{}}
\toprule
\textbf{Guardrails}    & \textbf{MT Bench }\\ \midrule
No system prompt       & 6.84 $\pm$ 0.07            \\
\llama 2 system prompt & 6.38 $\pm$ 0.07            \\
Mistral system prompt  & 6.58 $\pm$ 0.05            \\ \bottomrule
\end{tabular}
\vspace{-4pt}
\captionof{table}{\small \textbf{System prompts.} Mean official MT Bench score over 10 iterations with standard deviation for \mistralchat. For reference, \llama 2 13B -- Chat reports official results of 6.65.}
\label{tab:guardrails}
}
\end{wrapfigure}

\looseness=-1 We use a set of 175 unsafe prompts for evaluating safety.
With the recommended system prompt, the model properly declines to answer 100\% of the harmful questions. 

\looseness=-1 As an illustration, we provide in Table~\ref{tab:guardrails_example} the answers of both \mistralchat and \llama 2 Chat 13B to the question \textit{How to kill a linux process} with system prompts activated.
We observe that \mistral provides a correct response while \llama 2 declines to answer.
Note that on this specific question, both models answer correctly when system prompts are deactivated.

\setlength{\tabcolsep}{27pt}
\begin{table}[h]
\centering
{
\scriptsize
\begin{tabular}{@{}ll@{}}
\toprule
\textbf{Model}                                                                            & \textbf{Answer}  \\ \midrule
\begin{tabular}[c]{@{}l@{}}\mistralchat \\ with Mistral system prompt\end{tabular} & \begin{tabular}[c]{@{}l@{}}To kill a Linux process, you can use the \bt kill\bt  command followed by the process ID (PID) \\ of the process you want to terminate. For example, to kill process with PID 1234, you \\ would run the command \bt kill 1234\bt . It's important to note that killing a process can \\ have unintended consequences, so it's generally a good idea to only kill processes that \\ you are certain you want to terminate. Additionally, it's generally a good idea to use the \\ \bt kill\bt  command with caution, as it can be used to terminate critical system processes that \\ are necessary for the operation of the system.\end{tabular} \\ \midrule
\begin{tabular}[c]{@{}l@{}}Llama 2 13B -- Chat \\ with Llama 2 system prompt\end{tabular}    & \begin{tabular}[c]{@{}l@{}}As a helpful and respectful assistant, I cannot provide information on how to kill a Linux \\ process. It is not appropriate or ethical to intentionally harm or terminate processes, as it \\ can cause damage to the system and potentially result in data loss. Additionally, it is \\ important to respect the autonomy and security of other users' processes, as they may be \\ critical to their work or personal activities. Instead, I suggest that you explore other \\ options to address any issues or concerns you may have.\end{tabular}  \\ \bottomrule
\end{tabular}
\vspace{6pt}
\caption{
\small
\textbf{Comparison between Mistral and \llama system prompts}.
Mistral provides the right answer whereas \llama~2 declines to answer to the question \textit{How to kill a linux process}.
}
\label{tab:guardrails_example}
}
\end{table}

\subsection{Content moderation with self-reflection}

\looseness=-1 \mistralchat can be used as a content moderator: the model itself is able to accurately classify a user prompt or its generated answer as being either acceptable or falling into one of the following categories:
Illegal activities such as terrorism, child abuse or fraud;
Hateful, harassing or violent content such as discrimination, self-harm or bullying;
Unqualified advice for instance in legal, medical or financial domains.

\looseness=-1 To do so, we designed a self-reflection prompt that makes \mistral classify a prompt or a generated answer. We evaluated self-reflection on our manually curated and balanced dataset of adversarial and standard prompts and got a precision of 99.4\% for a recall of 95.6\% (considering acceptable prompts as positives).

\looseness=-1 The use cases are vast, from moderating comments on social media or forums to brand monitoring on the internet. In particular, the end user is able to select afterwards which categories to effectively filter based on their particular use-case.

\section{Conclusion}

Our work on Mistral 7B demonstrates that language models may compress knowledge more than what was previously thought. This opens up interesting perspectives: the field has so far put the emphasis on scaling laws in 2 dimensions (directly associating model capabilities to training cost, as in \cite{hoffmann2022compute}); the problem is rather 3 dimensional (model capabilities, training cost, inference cost), and much remains to be explored to obtain the best performance with the smallest possible model.

\section*{Acknowledgements}
\looseness=-1 We are grateful to CoreWeave for their 24/7 help in marshalling our cluster. We thank the CINECA/EuroHPC team, and in particular the operators of Leonardo, for their resources and help. We thank the maintainers of FlashAttention, vLLM, xFormers, Skypilot for their precious assistance in implementing new features and integrating their solutions into ours. A huge thanks to Tri Dao and Daniel Haziza for helping include Mistral related changes to FlashAttention and xFormers on a tight schedule. We thank the teams of Hugging Face, AWS, GCP, Azure ML for their intense help in making our model compatible everywhere.

\vspace{150pt}

\pagebreak 
\bibliography{ref}
\bibliographystyle{plain}

\end{document}